\documentclass[sigconf]{acmart}
\renewcommand\footnotetextcopyrightpermission[1]{} % removes footnote with conference information in first column
\usepackage{booktabs} % For formal tables

\usepackage{blindtext}
\usepackage{graphicx}
\usepackage{epstopdf}
\usepackage{bm}
\usepackage{amsfonts,amssymb}
\usepackage{amsmath}
\usepackage{verbatim}
\usepackage{caption}
\usepackage{subfigure}

\usepackage{amsthm} % for theorems and definitions
\newtheorem{myDef}{Definition}

\newcommand{\reffig}[1]{Figure \ref{#1}}
\newcommand{\reftab}[1]{Table \ref{#1}}
\newcommand{\refalg}[1]{Algorithm \ref{#1}}

\usepackage{tikz}
\usetikzlibrary{arrows,graphs}

\usepackage{algorithm}
\usepackage{algorithmic}

\usepackage{enumerate}

\usepackage{graphicx}
\usepackage{multirow}
\usepackage{booktabs}
\usepackage{makecell}

\setcopyright{none}
\begin{document}
\title{Diffusion Based Network Embedding}

\author{Yong Shi}
\affiliation{%
  \institution{School of Economics and Management, \\ University of Chinese Academy of Sciences}
  \city{Beijing}
  \country{China}
  \postcode{100190}
}
\email{yshi@ucas.ac.cn}

\author{Minglong Lei}
\affiliation{%
  \institution{School of Computer and Control Engineering, \\ University of Chinese Academy of Sciences}
  \city{Beijing}
  \country{China}
  \postcode{100049}
}
\email{leiminglong16@mails.ucas.ac.cn}

\author{Peng Zhang}
\affiliation{%
  \institution{Ant Financial}
  \city{Hangzhou}
  \country{China}}
\email{zhangpeng04@gmail.com}

\author{Lingfeng Niu}
\affiliation{%
  \institution{School of Economics and Management, \\ University of Chinese Academy of Sciences}
  \city{Beijing}
  \country{China}
  \postcode{100190}
}
\email{niulf@ucas.ac.cn}

\begin{abstract}
In network embedding, random walks play a fundamental role in preserving network structures. However, random walk based embedding methods have two limitations. First, random walk methods are fragile when the sampling frequency or the number of node sequences changes. Second, in disequilibrium networks such as highly biases networks, random walk methods often perform poorly due to the lack of global network information. In order to solve the limitations, we propose in this paper a network diffusion based embedding method. To solve the first limitation, our method employs a diffusion driven process to capture both depth information and breadth information. The time dimension is also attached to node sequences that can strengthen information preserving. To solve the second limitation, our method uses the network inference technique based on cascades to capture the global network information. To verify the performance, we conduct experiments on node classification tasks using the learned representations. Results show that compared with random walk based methods, diffusion based models are more robust when samplings under each node is rare. We also conduct experiments on a highly imbalanced network. Results shows that the proposed model are more robust under the biased network structure.
\end{abstract}


\keywords{Network Embedding, Cascades, Diffusion, Network Inference, Dimension Reduction.}

\maketitle
\section{Introduction}

Network representation learning\cite{yan2007graph, chang2015heterogeneous} has been widely used in large network analysis. The basic idea is to map graph nodes in the original feature space into a low-dimensional space while maintaining the network proximities and structure information\cite{wang2016structural,ribeiro2017struc2vec}. To date, network embedding has shown its advantage in improving the performance of network classification\cite{lu2003link}, anomaly detection\cite{akoglu2010oddball} and community detection\cite{rohe2011spectral}.

Early network embedding methods treat the learning problem as a dimension reduction problem and analyze the adjacent matrices and their variations. However, these methods fall into the category of deterministic models that can only handle static network connections.

Recently, graph sampling\cite{wang2011understanding,de2010does} has been widely used for embedding, where random walks are introduced to preserve network structures. Random walks closely related to the spectrum of networks\cite{lovasz1993random}. First, random walks are used to sample graphs, e.g., node2vec~\cite{grover2016node2vec} makes a tradeoff between breadth-first sampling (BFS) and depth-first sampling (DFS). Second, node sequences produced by the sampling process are fit into the skip-gram\cite{mikolov2013efficient} model that can encode the sampling results into a low-dimensional latent space.

However, the above random walk methods have two limitations. First, DFS and BFS are often not informative enough to capture the network structures. In random walks, the success of preserving network structure information highly depends on the repeated sampling imposed on each node. Therefore, these methods are fragile when the sampling frequency or the number of node sequences changes. Moreover, encoding local structure node sequences to low-dimensional representations by skip-gram\cite{mikolov2013efficient} is an end-to-end process. In disequilibrium networks such as highly biases networks\cite{gjoka2010walking}, the methods often perform poorly due to the lack of global network information.

To solve the above two shortcomings, we propose in this paper a diffusion based embedding model. Diffusion models can dynamically detect network structures and have been successfully used in dynamic network analysis\cite{gomez2010inferring,rodriguez2011uncovering,myers2010convexity}.
Specifically, the diffusion embedding method can be taken as a two-step framework which consists of a \emph{detecting step} and a \emph{mapping step}.

In the \emph{detecting step}, our method simulates the information diffusion process and generates a collection of node sequences. Unlike random walk based methods, our method remember all the visited nodes when running the algorithm. Such a modification transforms single-trace random walks into multiple-trace random walks. Without turning the parameters between BFS and DFS, the diffusion provides an intuitive way to detect both structures. Since more nodes are involved in the process of one sampling, our method is able to capture more local information than traditional random walk methods given the same walk length and sampling frequency. Another improvement over random walk is that we add an additional time dimension over the sequences. We argue that, besides BFS and DFS, an additional time dimension can make the description of local structures more comprehensive. We then formulate the diffusion cascades~\cite{leskovec2007patterns} with time information under pure node sequences.

In the \emph{mapping step}, based on the diffusion cascades, we infer the network by network inference technique\cite{rodriguez2011uncovering} and obtain a weight matrix which describes the network connections. Instead of directly encoding the sampling sequences, network inference is launched over the whole network. The weight matrix is consequently able to capture the global structure information. Diffusion embedding is then more robust to the unbalance structures of network. At the last step, we apply a simple SVD factorization as the dimension reduction method to the weight matrix to obtain the low-dimensional representation.

We conduct extensive experiments in the node classification task to evaluate the proposed method. Comparing with baseline methods, our method achieves better performance, which indicates that the learned representations can better reveal the network structure.

The contributions of this paper can be summarized as follows:

(1) We propose a novel network embedding method based on information diffusion in networks. Unlike random walk based methods, our method remember all the visited nodes. Such a modification transforms single-trace random walks into multiple-trace random walks.

(2) We propose a new strategy to capture the structural information. Given the same walk length and sampling frequency, our method is able to capture more local information than traditional random walk methods.

(3) We conduct extensive experiments on real datasets for node classification. The results shows that our method outperforms baselines.

\section{Related Work}

\subsection{Network Embedding Methods}

Network embedding is a subtopic of representation learning in networks. Early methods such as Laplacian Eigenmaps(LE)\cite{belkin2001laplacian}, Local Linear Embedding(LLE)\cite{roweis2000nonlinear} and IsoMAP\cite{tenenbaum2000global} are served as dimension reduction techniques which are not originally designed for networks. In these methods, graphs are constructed off-line by computing the distances of node attributes. The representations are obtained by solving the eigenvectors of adjacent matrices or their variations.

Generally, embedding real world networks demands models to preserve actual graph proximity in learned features. Except for dimension reduction\cite{yan2007graph}, another intuitive solution is utilizing matrix factorization \cite{ahmed2013distributed, cao2015grarep} to get the low-dimensional representations. Similar to previous described dimension reduction methods, the graph proximities are revealed directly by adjacent matrices and their variations. One basic assumption in this category of methods is that the connection between any two vertices is denoted as dot product of their low-dimensional embeddings\cite{hamilton2017representation}. In a common sense, matrix based methods are making trade-off decisions among different orders of structural information. Since they mainly depends on visible and deterministic connections of vertices, the latent structures in real networks can not be properly exploited.

Recently, graph sampling based methods have achieved significant success in network embedding. Instead of measuring the proximity of graph by deterministic edges, sampling methods capture the vertex proximity by stochastic measure\cite{hamilton2017representation}. Vertices that appeared in a same vertex sequence are supposed to have similar representations. DeepWalk\cite{perozzi2014deepwalk} and node2vec \cite{grover2016node2vec} adopt different sampling strategies to sample local based sequences. Those local vertex sequences are directly decoded into latent representation space and hence fail to capture the global information. Another work, LINE\cite{tang2015line}, is a large-scale information embedding method which designs a loss function that captures both 1-step and 2-step proximity information. LINE is also treated as a stochastic embedding model since it optimizes a probabilistic loss function. However, LINE still suffers from losing the presentation of global information.

Our proposed method in this paper has overcome several disadvantages of above models. We adopt stochastic sampling to keep local structures of networks comparing with conventional matrix based models. Unlike random walk driven models, the sequences produced by diffusion sampling is much informative both in the node and time perspective. The proposed model also provides a global picture of the network.

\subsection{Random Walk and Diffusion}

In this part, we briefly introduce the basic ideas of random walk and diffusion process. Random walk and diffusion are originally studied in physics that describe the molecule movements. The general random walk refers to a discrete stochastic process\cite{itoandh1965diffusion}. For example, a simplest random walk is defined over an integer sequence line with a probability of $1/2$ going right with $\triangle x=1$ and a probability of $1/2$ going left with $\triangle x=-1$. Denote the site that the walker is at in time step $t$ as $X_{t}$, then
\begin{equation}\label{eq1}
 X_{t}=X_{t-1}+\triangle x
\end{equation}

However, the diffusion is defined with continuous space and continuous time by a stochastic differential equation in the following form\cite{itoandh1965diffusion}:
\begin{equation}\label{eq2}
  dX_{t}=\mu(X_{t})dt+\delta(X_{t})dW_{t}
\end{equation}
where $W$ describes the Brownian motion which is highly related to random walk models, $\delta$ is the diffusion coefficient and $\mu$ is the drift term.

It is obvious that diffusion possesses randomness of random walk and are subject to more sophisticated stochastic rules\cite{itoandh1965diffusion}. In the network research, the random walk is a basic block of diffusion process such as epidemic spreading\cite{newman2002spread} and opinions propagatation\cite{watts2007influentials}.

\section{The Model}

\subsection{The Diffusion Process}

It has been practically proved that random walk is a powerful tool to traversal the network structure with a collection of node sequences\cite{perozzi2014deepwalk}. However, the diffusion over network generates more informative traces that not only consist node sequences but also information fragments along with the nodes. In this subsection, we illustrate how the diffusion process can be utilized as an efficient tool to detect the network structure.

Firstly, we briefly explain how the diffusion happens under a network. Exactly as the molecular diffusion in fluid where particles move from high concentration area to low concentration area, the network can be regarded as a system that changes from an unstable state to a stable state throughout the diffusion process\cite{abrahamson1997social}. Initially, the network is an unbalanced system where only a few nodes are active. Since there are information gaps between different nodes, the diffusion happens when information is delivered from active nodes to inactive ones. The system will be stable when the information is evenly distributed.

For convenience, we define the graph under a network as follows:

\begin{myDef}(Graph)
Given $N$ vertices, a graph can be defined as a tuple $G(V,E)$ where $V$ is the vertex set denoted as $\{v_{1},\cdots,v_{n}\}$ and $E$ is the edges set denoted as $\{v_{i,j}\}_{i,j=1}^{n}$.
\end{myDef}

Unlike single-trace random walks, the node sampling process of diffusion under $G(V,E)$ generates a series of node sets that describing the evolution of the participated nodes. Concretely, let us choose a random vertex $v_{i}(v_{i}\in V)$ as the seed to start a diffusion process. Supposing the maximal walking step is $K$. Given an arbitrary step $k$, $S_{v_{i}}^{k}$ denotes a subset of $V$ which includes all vertices that are active in the current step. In step $k+1$, all vertices in $S_{v_{i}}^{k}$ are served as seeds to launch node samplings. We use $D_{v_{i}}^{k+1}$ to denote the vertex set that generated from $S_{v_{i}}^{k}$. Each vertex in $D_{v_{i}}^{k+1}$ is randomly selected from the neighbors of a corresponding vertex in $S_{v_{i}}^{k}$. Then $S_{v_{i}}^{k}$ can update to $S_{v_{i}}^{k+1}$ by adding new infected nodes in step $k$.
Therefore, for each vertex $v_{i}$, we obtain  $D_{v_{i}}:=(D_{v_{i}}^{0},D_{v_{i}}^{1},\cdots,D_{v_{i}}^{K})$ and $S_{v_{i}}:=(S_{v_{i}}^{0},S_{v_{i}}^{1},\cdots,S_{v_{i}}^{K})$, where $K$ is the length of walking steps.

\begin{figure*}
  \centering
  \subfigure[Generate a random walk sequence]{
    \label{fig1:subfig:a}
    \includegraphics[scale=0.34]{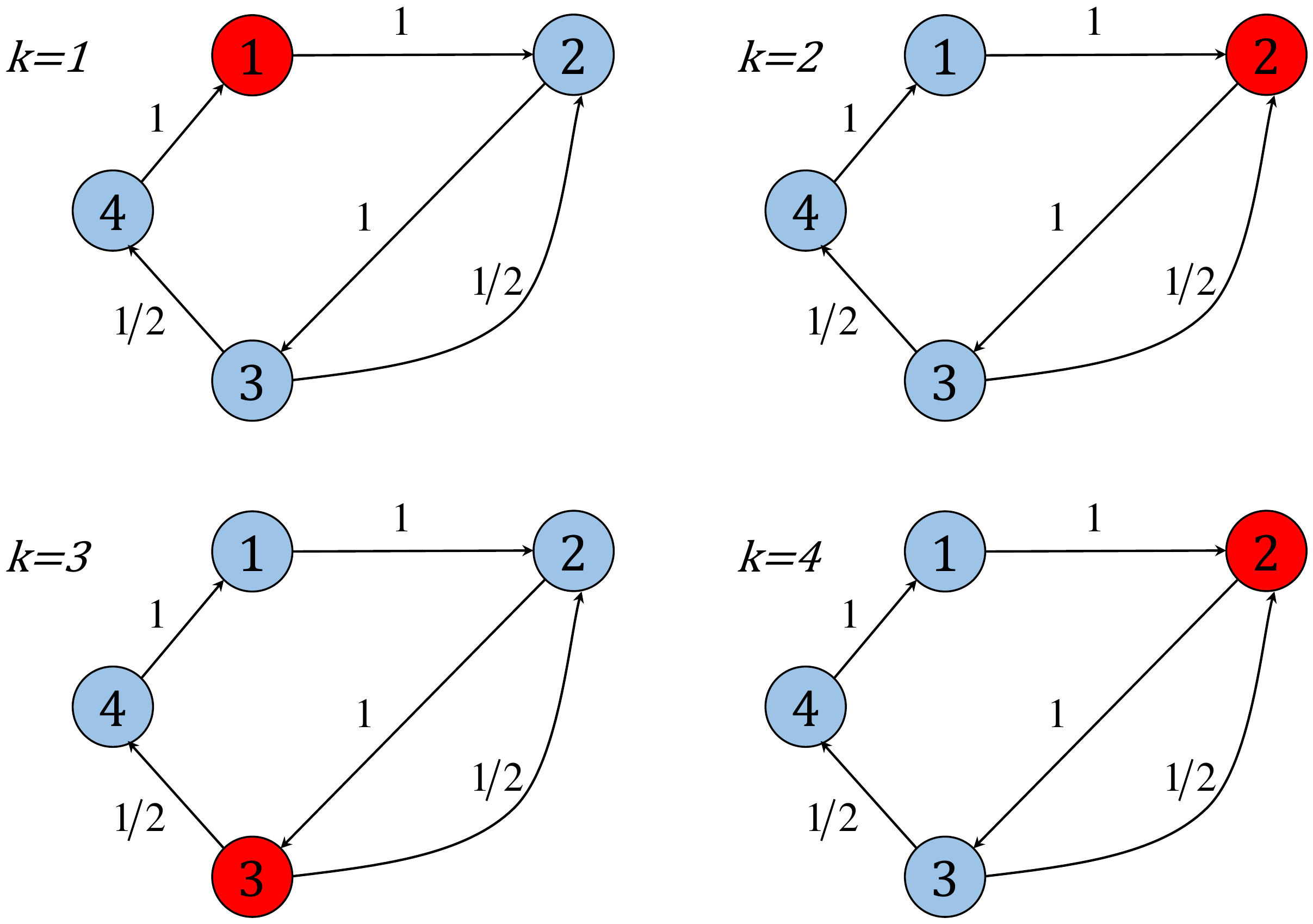}}
  \hspace{0.2in}
  \subfigure[Generate a diffusion sequence]{
    \label{fig1:subfig:b}
    \centering
    \includegraphics[scale=0.34]{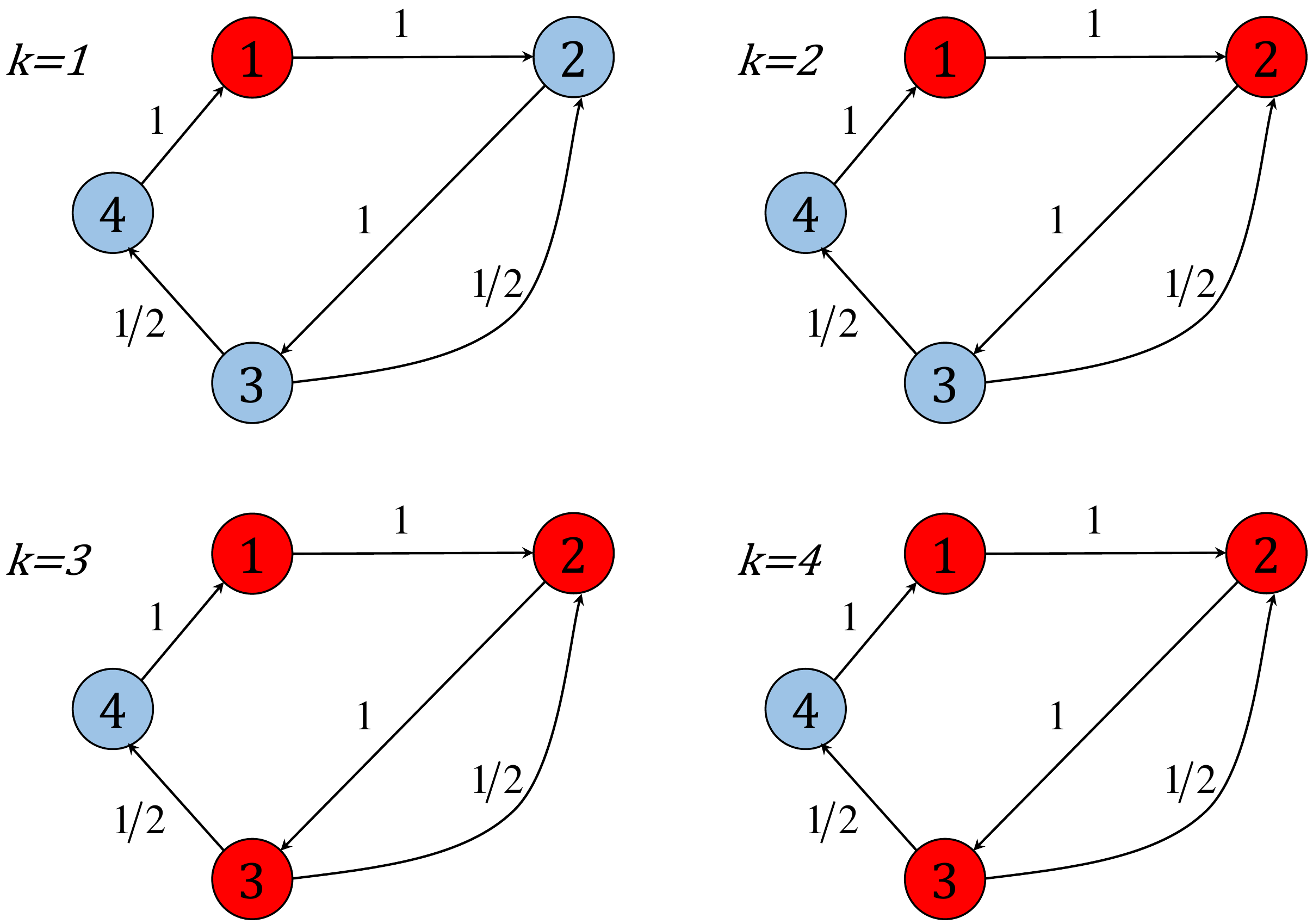}}
  \caption{A simple example of random walk and diffusion process in a directed graph.}
  \label{fig1:subfig}
\end{figure*}

We compare tradition random walks and our diffusion process in a directed graph with four nodes in \reffig{fig1:subfig}. The walking step $K$ is set to be four for the convenience of illustration. In \reffig{fig1:subfig:a}, we firstly launch a random walk start at node $v_{1}$. In each walking step, the walker moves to its neighbors with uniform probabilities. The red circle indicates the current node that the walker is at. After a four-step walk, it generates a node sequence $(v_{1},v_{2},v_{3},v_{2})$ which reflects the local network structure. In \reffig{fig1:subfig:b}, we simulate a diffusion process under a directed graph also starts at node $v_{1}$.
The difference between the two kinds of random samplings is that the diffusion process is memorable since an active node will stay active after it has been visited. The diffusion process generates a sequences of node sets$((v_{1}),(v_{1},v_{2}),(v_{1},v_{2},v_{3}),(v_{1},v_{2},v_{3}))$. As illustrated in \reffig{fig1:subfig}, if we record a node sequence within $K=5$ steps, the node $v_{4}$ in a random walk sequence will not be visited since the walker has passed through all neighbors of $v_{4}$. However, since the diffusion walk stores all nodes that have been visited previously, $v_{1}$,$v_{2}$,$v_{3}$ will stay active at the same time, the diffusion walker is possible to visit $v_{4}$ when $k=5$.

Obviously, each time when we start a sampling during a diffusion process, different from random walk in which only one node is working as seed, an already active node can sample multiple times in order to detect more structural information. The walker is then improved from a single-sampling trace to a multiple-sampling trace where there are more possibilities to discover local structures of networks.

\subsection{The Formulation of Cascades}

The simulation of diffusion under a network is step discrete and time continuous. The discrete aspect of diffusion, as illustrated in the previous subsection, derives the ability from random walk to perceive local structures of networks. In contrast, the continuous aspect of diffusion provides us more information about the latent global structure of the network.

Generally, the transmission of knowledge or disease is not instantly happened in the diffusion process\cite{kempe2003maximizing}. Actually, another important feature in a diffusion process that also contributes to discover the structural information is the $timing$ that a vertex receives the information. For example, in the epidemiology scenario\cite{newman2002spread}, it indicates the $timing$ when vertices get infected by a disease. We utilize the information $cascades$\cite{leskovec2007patterns} to record the infected time of nodes in the process of diffusion.

Models based on cascades have been applied in many circumstances such as recommendation systems\cite{leskovec2006patterns,leskovec2007dynamics}. For many works that aim to discover network structures, cascades record the flows of information in networks\cite{rodriguez2011uncovering}. Considering a graph with $N$ vertices. If we fix an observation window $[0,T^{c}]$ in advance, a cascade is defined as a $N$-dimensional vector $\mathbf{t}^{c}:=(t^{c}_{1},\cdots,t^{c}_{N})$ where each element of the vector records the first infection time of the corresponding node. In other words, for any $k\in[1,N]$, $t^{c}_{k} \in [0,T^{c}] \cup \{\infty\}$. Here $\infty$ implies that the node is not observed to be infected within the observation window. Each nodes in one cascade has been attached with a timestamp. Then a collection of $|C|$ cascades can be represented as $\{\mathbf{t}^{1},\cdots,\mathbf{t}^{|C|}\}$. Note that each time a cascade is generated, the timing will be reset to 0. For simplicity, the observation windows are set to be equal-timing: $T^{c} = T, for\ c \in [1,|C|]$.

Recall that we have obtained a collection of node sets $S_{v_{i}}$ for vertex $v_{i}$ that is generated by node sampling process. In order to formulate cascades by previous node sets, we introduce the time interval sampling strategy. The transmission time between two nodes can be depicted by proper transmission time models\cite{myers2010convexity}. The most common models in present are power law model in which the time interval $t$ subjects to $d(t)=(\alpha-1) t^{(-\alpha)}$ and exponential model in which the time interval $t$ subjects to $d(t)=\alpha e^{-\alpha t}$. The time interval samplings are launched along with the node samplings. In detail, for each node that is active in the current step, we sample a neighbor of the current node as its next stop and then sample a time interval $t$ from time models as the transmission time of diffusion. The timestamp of $v_{i}$ is set to be $0$ at the beginning. As the walking proceed, newly infected nodes from $S_{v_{i}}^{k}$ to $S_{v_{i}}^{k+1}$ will be assigned with timestamps based on the time intervals and their source of infections. Notice that an already active node cannot be infected even the walker can revisit it. Consequently, we only record the time of its first infection. After a $K$ step walking process, $S_{v_{i}}^{K}$ records all nodes that are involved in a diffusion process with timestamps . Then, we can select nodes in $S_{v_{i}}^{K}$ that within observation window $T^{c}$ to formulate a cascade $\mathbf{t}^{v_{i}}:=(t^{v_{i}}_{1},\cdots,t^{v_{i}}_{N})$. The cascades collection $\{\mathbf{t}^{1},\cdots,\mathbf{t}^{|C|}\}$ are then obtained by conducting samplings in different nodes.

Also recall that the update of $S_{v_{i}}$ is the evolution of the network. The joining of new nodes and edges at each step implies changeable dynamic process of networks. After incorporating an additional dimension of information into node sequences, the cascades derive temporal features, which is valuable for detecting the underlying structural information. Consider a cascade $\mathbf{t}^{c}$, the timestamps depict orders of nodes implicitly by their values. In this way, the timesamps give us global positions of involved nodes during a diffusion process. Consequently, a certain amount of cascades can be used together to inference the global structures of a network.

\subsection{Network Inference}

After obtaining $|C|$ cascades, we want to transform them into a more accurate network representation that contains both global and local structural information. This aim can be fulfilled by solving a specific optimization problem based on the network inference method\cite{rodriguez2011uncovering}.

We start with the definition of the pairwise transformation likelihood of any two nodes. For an arbitrary node $v_{j}$, the probability that it is infected by an active node $v_{i}$ is defined as a likelihood function in the format $f(t_{v_{j}}|t_{v_{i}}, \alpha_{v_{i},v_{j}})$. The $\alpha_{v_{i},v_{j}}$ is the transformation rate between two nodes and $t_{v_{i}},t_{v_{j}}(t_{v_{i}}<t_{v_{j}})$ are the time of infection of node $v_{i}$ and node $v_{j}$ respectively. Take the exponential model for example, the parametric form of the conditional likelihood can be written as:

\begin{equation}\label{eq3}
\begin{cases}
\alpha_{v_{i},v_{j}} \cdot e^{-\alpha_{v_{i},v_{j}}(t_{v_{j}}-t_{v_{i}})} & {if \ t_{v_{i}}<t_{v_{j}}} \\
0 & {otherwise}
\end{cases}
\end{equation}
Furthermore, the cumulative density function $F(t_{v_{j}}|t_{v_{i}}, \alpha_{v_{i},v_{j}})$ is computed from $f(t_{v_{j}}|t_{v_{i}}, \alpha_{v_{i},v_{j}})$. Then the probability that node $v_{j}$ is not infected by an already infected node $v_{i}$ is defined as $S(t_{v_{j}}|t_{v_{i}}, \alpha_{v_{i},v_{j}}) = 1 - F(t_{v_{j}}|t_{v_{i}}, \alpha_{v_{i},v_{j}})$.

Hence, the likelihood of one given cascade is given by:
\begin{flalign}\label{eq4}
\begin{split}
f(\mathbf{t};\mathbf{A}) = \prod_{t_{v_{j}}\leq T} S(T|t_{v_{j}};\alpha_{v_{j},m}) \times \\
\prod_{k:t_{k}<t_{v_{j}}} S(T|t_{v_{j}};\alpha_{k,v_{j}}) \sum_{i:t_{v_{i}}<t_{v_{j}}} H(t_{v_{j}}|t_{v_{i}};\alpha_{v_{i},v_{j}})
\end{split}
\end{flalign}
where the $H(t_{v_{j}}|t_{v_{i}};\alpha_{v_{i},v_{j}}) = \frac{f(t_{v_{j}}|t_{v_{i}}, \alpha_{v_{i},v_{j}})}{S(t_{v_{j}}|t_{v_{i}}, \alpha_{v_{i},v_{j}})}$ is the instantaneous transmission rate from node $i$ to node $j$.

For a set of $|C|$ independent cascades, we define the joint likelihood as:
\begin{equation}\label{eq5}
\prod_{\mathbf{t}^{c} \in C}f(\mathbf{t}^{c};\mathbf{A})
\end{equation}

In all, by estimating the transmission rate $\alpha_{v_{i},v_{j}}$, the network inference problem is formulated with a loss function:
\begin{equation}\label{eq6}
\begin{split}
\mathop{\min}\limits_{\mathbf{A}} \quad & -\sum_{c \in C} \log f(\mathbf{t}^{c};\mathbf{A}) \\
s.t. \quad & \alpha_{v_{i},v_{j}} \geq 0, v_{i},v_{j} = 1,\cdots,N, v_{i} \neq j
\end{split}
\end{equation}

Denote the optimal solution of \eqref{eq6} as $\mathbf{A}$. It provides an accurate weight which reflects the degree of connection strongness between each pair of nodes. Since the inference is launched under the whole network, the resulting weight matrix $\mathbf{A}$ is able to capturing the global information of the network.

\subsection{Dimension Reduction}

In order to examine whether the diffusion process with a network inference is effective enough to preserve structural information of the network, we use a simple matrix factorization method, singular value deposition(SVD), to get the ultimate low-dimensional representations of the vertices. SVD has been successfully used as a dimensional reduction tool in \cite{levy2014neural}. It is also feasible to employ deep models as suggested in \cite{wang2016structural,tian2014learning,cao2016deep} to learn the embeddings.

Recall that we have inferenced a weight matrix $\mathbf{A}$ that combines both local and global information about the network. SVD factorizes the normalized matrix as:

\begin{equation}\label{eq7}
\mathbf{A} = U \Sigma (V)^{T}
\end{equation}
where $\Sigma$ is a diagonal matrix whose non-zero elements are singular values of $\mathbf{A}$, $U$ and $V$ are two orthogonal matrices, which are composed of the left and right singular vectors respectively.

Given the dimension of embedding vectors $d$, $\Sigma_{d}$ is an approximation of $\Sigma$ which only considers top-$d$ singular values of $\Sigma$. Follow the method in \cite{levy2014neural} and \cite{cao2015grarep}, let $\mathbf{A}_{d}$ with rank $d$ to be the approximation of $\mathbf{A}$, the approximation factorization of $\mathbf{A}$ can be written as:
\begin{equation}\label{eq8}
\mathbf{A} \approx \mathbf{A}_{d} = U_{d} \Sigma_{d} (V_{d})^{T} = \mathbf{Y} \mathbf{W}
\end{equation}
where $\mathbf{Y}=U_{d}(\Sigma_{d})^{\frac{1}{2}}$ and $\mathbf{W}=(\Sigma_{d})^{\frac{1}{2}}V_{d}^{T}$. Then $\textbf{Y}$ can be used as the final representation of the network.

The whole algorithm for completeness is listed in \refalg{alg1}.

\begin{algorithm}[!h]
  \caption{Diffusion Based Network Embedding}
    \label{alg1}
  \begin{algorithmic}[1]
    \REQUIRE graph $G(V,E)$\\
    diffusion steps $K$\\
    representation size $d$\\
    time window size $T^{c}$\\
    number of samplings under each vertex $\tau$
    \ENSURE
    $N$ cascades $\{\mathbf{t}^{1},\cdots,\mathbf{t}^{|C|}\}$ \\
    weight matrix $\mathbf{A}$ \\
    network representations in a matrix form $\mathbf{Y}$
    \FOR {$i=1$ to $\tau$}
      \STATE $\Omega = Shuffle(V)$
      \FOR {$v_{i} \in \Omega$}
      \STATE Run node sampling process($G,v_{i},K,\tau,T^{c}$) to obtain $D_{v_{i}},S_{v_{i}}$
      \STATE Run time sampling process($D_{v_{i}},S_{v_{i}}$) to obtain$S_{v_{i}}^{t}$
      \STATE Formulate a cascade $\mathbf{t}^{v_{i}}$ within the time window $T^{c}$
      \ENDFOR
    \ENDFOR
    \STATE Solve Network inference optimization problem in \eqref{eq6} to obtain the transmission matrix $\mathbf{A}$
    \STATE Compute SVD optimization of $\mathbf{A}$ in \eqref{eq8} to obtain the final representations $\mathbf{Y}$
  \end{algorithmic}
\end{algorithm}

\section{Experiments}

In this section, we conduct experiments on different real network datasets in node classification tasks to testify the effectiveness of the proposed method.

\subsection{Datasets}

\begin{enumerate}[1.]
  \item \textbf{Wiki Network.} \cite{yang2015network} Wiki network is a collection of 2,405 websites from 17 categories. There are 17,981 links between them. The edge between every two pages indicate a hyperlink.
  \item \textbf{Citeseer Network}\cite{lu2003link, sen2008collective}. Citeseer\footnote{http://citeseer.ist.psu.edu/index} is a popular scientific digital library and academic search engine. Citeseer network is a document based citation network that consists of 3312 scientific papers. All of those publications can be classified into 6 different groups: Agents, Artificial Intelligence, Database, Human Interaction, Machine Learning and Information Retrieval. This network consists of 4732 links that describe the citation relations between different papers. The papers in the dataset are selected so that each paper will cite or be cited by at least one paper.
  \item \textbf{Cora Network} \cite{lu2003link, sen2008collective}. The Cora network is a citation network that consists of 2708 machine learning publications classified into one of 7 classes. This network consists of 5429 links indicating the citation relations between papers.
  \item \textbf{DBLP Network}\cite{sun2009ranking}. The DBLP dataset is a co-authorship network that concentrates on computer science publications. The original DBLP dataset in \cite{yang2015defining} is used for community evaluation. Following the experiment settings in \cite{sun2009ranking}, we construct a four-area dataset which contains Machine Learning, Data Mining, Information Retrieval and Database. For each area, five representative conferences are selected. Authors that published papers in those conferences are selected and constructed as the author network.
\end{enumerate}

It is noticeable that cora network originally used in \cite{lu2003link} includes both attribute information and connection information. In this paper, we remove the attribute information and only consider the 0-1 link information. The statics of the datasets are summarized in \reftab{table1}.

\begin{table}
  \small
  \centering
  %\captionsetup{font={footnotesize}}
  \caption{Statics of datasets}
  \begin{tabular}{*{4}{c}}
  \Xhline{2\arrayrulewidth}
  Dataset & $|V|$ & $|E|$ & $|Label|$ \\
  \Xhline{2\arrayrulewidth}
  WIKI     & 2405 & 17981 & 17 \\
  CORA     & 2708 & 5429  & 7  \\
  CITESEER & 3312 & 4732  & 6  \\
  DBLP     & 2760 & 3818  & 4  \\
  \Xhline{2\arrayrulewidth}
  \end{tabular}
  \label{table1}
\end{table}

\subsection{Baseline Methods}

To evaluate the performance of our method, several state-of-the-art network embedding methods are utilized as baselines.
 \begin{enumerate}[1.]
 \item \textbf{DeepWalk}\cite{perozzi2014deepwalk}. DeepWalk is a random walk based method that aims to learn low dimensional representations for networks. Random walk is served as network structure detector to obtain a collection of node sequences. The skip-gram is then adopted to obtain the final vector-wise representations.
 \item \textbf{GraRep}\cite{cao2015grarep}. GrapRep is a matrix factorization based graph representation method which utilizes $k$-step information matrix of the graph. For each single step information, GraRep learns a low dimensional embedding by imposing SVD on matrix that indicates the given step information. The final representations with global information is concatenated from all $k$ steps embeddings.
 \item \textbf{LINE}\cite{tang2015line}. LINE is designed for large scale information networks which attempts to preserve first-order structure information and second-order structure information by an explicit loss function. The final representation is concatenated from the first-order information and second-order information.
 \item \textbf{Spectral Clustering}\cite{ng2002spectral}.Spectral clustering operates on the Laplacian matrix of graph $G$ and can be utilized as an dimensional reduction method. The $d$ dimensional representations are generated from the top $d$ eigenvectors of normalized Laplacian matrix.
 \end{enumerate}

\begin{table*}
  \small
  \centering
  %\captionsetup{font={footnotesize}}
  \caption{Results on Wiki}
  \begin{tabular}{c|l|*{8}{p{1cm}<{\centering}|}p{1cm}<{\centering}}
  \Xhline{2\arrayrulewidth}
  Measure & Methods & 10\% & 20\% & 30\% & 40\% & 50\% & 60\% & 70\% & 80\% & 90\% \\
  \Xhline{2\arrayrulewidth}
  \multirow{5}{*}{Micro-F1}
  &Diffusion         & 0.5501 & \textbf{0.6080} & \textbf{0.6186} & \textbf{0.6285} & \textbf{0.6417} & \textbf{0.6578} & \textbf{0.6590} & \textbf{0.6715} & \textbf{0.6763} \\
  &DeepWalk          & 0.4997 & 0.5456 & 0.5801 & 0.6036 & 0.6234 & 0.6379 & 0.6445 & 0.6448 & 0.6453 \\
  &GraRep            & 0.5206 & 0.5972 & 0.6087 & 0.6138 & 0.6261 & 0.6292 & 0.6395 & 0.6501 & 0.6543 \\
  &LINE              & \textbf{0.5782} & 0.5902 & 0.6026 & 0.6089 & 0.6119 & 0.6243 & 0.6321 & 0.6376 & 0.6490 \\
  &SpetralClustering & 0.5452 & 0.6020 & 0.6052 & 0.6281 & 0.6396 & 0.6444 & 0.6515 & 0.6550 & 0.6515 \\
  \hline
  \multirow{5}{*}{Macro-F1}
  &Diffusion         & 0.3806 & 0.4183 &  0.4542 & \textbf{0.4794} & \textbf{0.4954} & \textbf{0.5082} & \textbf{0.5158} & \textbf{0.5227} & \textbf{0.5491}    \\
  &DeepWalk          & 0.3815 & 0.4231 & 0.4515 & 0.4760 & 0.4865 & 0.5060 & 0.5128 & 0.5198 & 0.5454 \\
  &GraRep            & \textbf{0.4016} & \textbf{0.4356} & 0.4526 & 0.4654 & 0.4605 & 0.4782 & 0.4884 & 0.4933 & 0.5042 \\
  &LINE              & 0.3838 & 0.4042 & 0.4228 & 0.4302 & 0.4358 & 0.4434 & 0.4530 & 0.4636 & 0.4873 \\
  &SpetralClustering & 0.3631 & 0.4320 & \textbf{0.4611} & 0.4789 & 0.4931 & 0.5047 & 0.5090 & 0.5160 & 0.5230 \\
  \Xhline{2\arrayrulewidth}
  \end{tabular}
  \label{table2}
\end{table*}

\begin{table*}
  \small
  \centering
  %\captionsetup{font={footnotesize}}
  \caption{Results on Citeseer}
  \begin{tabular}{c|l|*{8}{p{1cm}<{\centering}|}p{1cm}<{\centering}}
  \Xhline{2\arrayrulewidth}
  Measure & Methods & 10\% & 20\% & 30\% & 40\% & 50\% & 60\% & 70\% & 80\% & 90\% \\
  \Xhline{2\arrayrulewidth}
  \multirow{5}{*}{Micro-F1}
  &Diffusion         & \textbf{0.5042} & \textbf{0.5392} & \textbf{0.5640} & \textbf{0.5709} & \textbf{0.5899} & \textbf{0.5954} & 0.6076 & \textbf{0.6214} & \textbf{0.6355} \\
  &DeepWalk          & 0.4880 & 0.5350 & 0.5470 & 0.5587 & 0.5660 & 0.5702 & 0.5708 & 0.5741 & 0.5756 \\
  &GraRep            & 0.4116 & 0.4768 & 0.5257 & 0.5558 & 0.5762 & 0.5904 & 0.6038 & 0.6120 & 0.6231 \\
  &LINE              & 0.5002 & 0.5116 & 0.5231 & 0.5495 & 0.5780 & 0.5879 & 0.5917 & 0.5923 & 0.5996 \\
  &SpetralClustering & 0.4079 & 0.4865 & 0.5282 & 0.5571 & 0.5798 & 0.5914 & \textbf{0.6143} & 0.6195 & 0.6238 \\
  \hline
  \multirow{5}{*}{Macro-F1}
  &Diffusion         & \textbf{0.4439} & \textbf{0.4873} &  \textbf{0.5044} & \textbf{0.5088} & \textbf{0.5407} & \textbf{0.5578} & 0.5643 & \textbf{0.5728} & \textbf{0.5804}    \\
  &DeepWalk          & 0.4107 & 0.4452 & 0.4677 & 0.4808 & 0.4930 & 0.5192 & 0.5251 & 0.5281 & 0.5241 \\
  &GraRep            & 0.3521 & 0.4290 & 0.4780 & 0.5082 & 0.5277 & 0.5436 & 0.5595 & 0.5665 & 0.5748 \\
  &LINE              & 0.4253 & 0.4652 & 0.4734 & 0.4793 & 0.4921 & 0.5331 & 0.5515 & 0.5534 & 0.5557 \\
  &SpetralClustering & 0.3540 & 0.4358 & 0.4791 & 0.5085 & 0.5338 & 0.5525 & \textbf{0.5692} & 0.5711 & 0.5770 \\
  \Xhline{2\arrayrulewidth}
  \end{tabular}
  \label{table3}
\end{table*}

\begin{table*}
  \small
  \centering
  %\captionsetup{font={footnotesize}}
  \caption{Results on Cora}
  \begin{tabular}{c|l|*{8}{p{1cm}<{\centering}|}p{1cm}<{\centering}}
  \Xhline{2\arrayrulewidth}
  Measure & Methods & 10\% & 20\% & 30\% & 40\% & 50\% & 60\% & 70\% & 80\% & 90\% \\
  \Xhline{2\arrayrulewidth}
  \multirow{5}{*}{Micro-F1}
  &Diffusion         & 0.6554 & \textbf{0.7476} & \textbf{0.7695} & \textbf{0.7784} & \textbf{0.7854} & \textbf{0.7934} & \textbf{0.8018} & \textbf{0.8110} & \textbf{0.8128} \\
  &DeepWalk          & \textbf{0.6886} & 0.7243 & 0.7459 & 0.7600 & 0.7684 & 0.7675 & 0.7817 & 0.8059 & 0.8084 \\
  &GraRep            & 0.5217 & 0.6023 & 0.6497 & 0.6841 & 0.7086 & 0.7288 & 0.7415 & 0.7418 & 0.7413 \\
  &LINE              & 0.6763 & 0.7120 & 0.7452 & 0.7501 & 0.7673 & 0.7699 & 0.7732 & 0.7767 & 0.7780 \\
  &SpetralClustering & 0.5218 & 0.6101 & 0.6585 & 0.6862 & 0.7058 & 0.7164 & 0.7271 & 0.7335 & 0.7417 \\
  \hline
  \multirow{5}{*}{Macro-F1}
  &Diffusion         & 0.6270 & \textbf{0.7338} & \textbf{0.7510} & \textbf{0.7659} & \textbf{0.7668} & \textbf{0.7775} & \textbf{0.7954} & \textbf{0.8012} & \textbf{0.8086}    \\
  &DeepWalk          & \textbf{0.6683} & 0.7085 & 0.7302 & 0.7447 & 0.7536 & 0.7686 & 0.7718 & 0.7962 & 0.7996 \\
  &GraRep            & 0.4647 & 0.5765 & 0.6349 & 0.6712 & 0.6981 & 0.7219 & 0.7333 & 0.7361 & 0.7370 \\
  &LINE              & 0.6579 & 0.6840 & 0.7162 & 0.7262 & 0.7293 & 0.7368 & 0.7494 & 0.7503 & 0.7506 \\
  &SpetralClustering & 0.4643 & 0.5869 & 0.6452 & 0.6768 & 0.6975 & 0.7082 & 0.7166 & 0.7254 & 0.7423 \\
  \Xhline{2\arrayrulewidth}
  \end{tabular}
  \label{table4}

\end{table*}

\begin{figure}

%\begin{tabular}{cc}

\begin{minipage}{0.5\textwidth}
  \scshape
  \footnotesize

  \centerline{\includegraphics[width=1\textwidth,trim=60 0 60 0,clip]{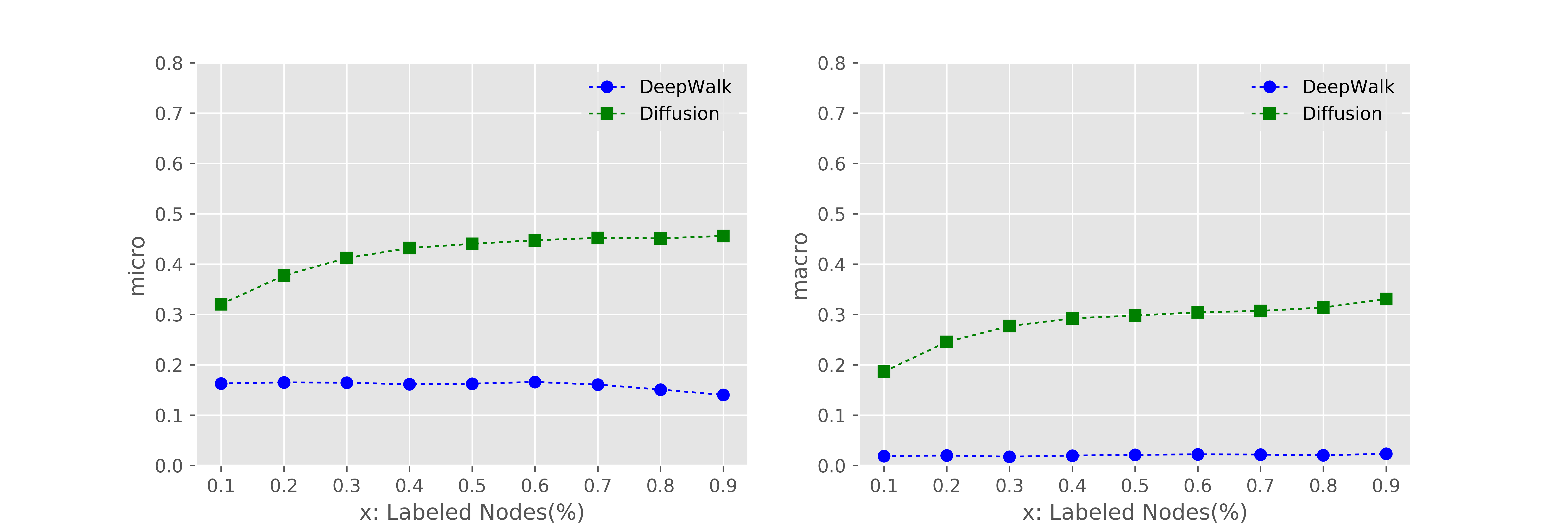}}

  \centerline{(a) Wiki, $T^{c}=10(T^{w}=10)$, $\tau=1(\gamma=1)$}

\end{minipage}

\hfill

\begin{minipage}{0.5\textwidth}
  \scshape
  \footnotesize

  \centerline{\includegraphics[width=1\textwidth,trim=60 0 60 0,clip]{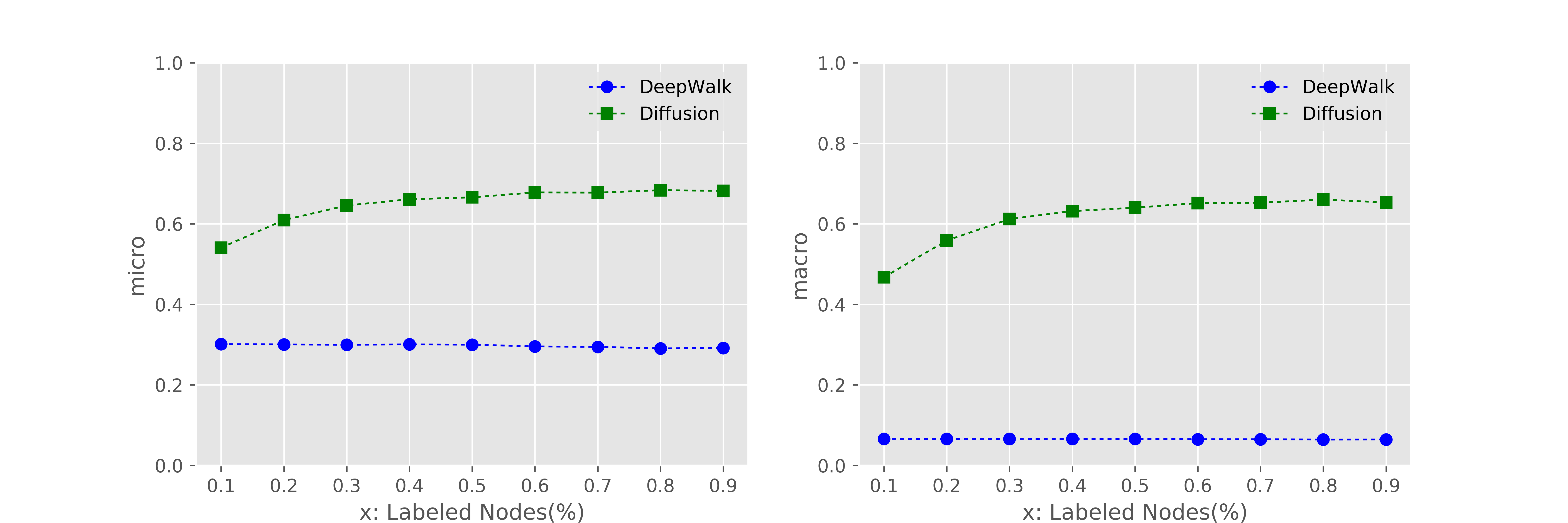}}

  \centerline{(b) Cora, $T^{c}=10(T^{w}=10)$, $\tau=1(\gamma=1)$}

\end{minipage}

\vfill

\begin{minipage}{0.5\textwidth}
  \scshape
  \footnotesize

  \centerline{\includegraphics[width=1\textwidth,trim=60 0 60 0,clip]{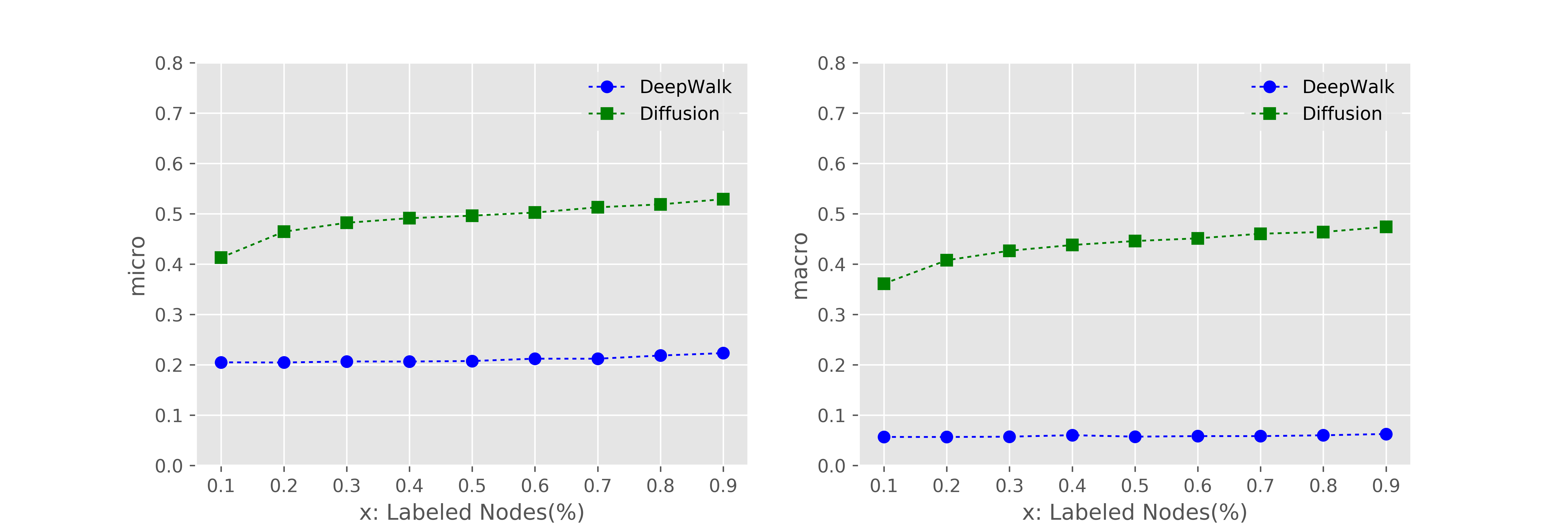}}

  \centerline{(c) Citeseer, $T^{c}=10(T^{w}=10)$, $\tau=1(\gamma=1)$}

\end{minipage}

%\end{tabular}

\caption{An extreme example when the number of samplings on each node is 1.}

\label{fig:figure_2}

\end{figure}

\begin{figure}

%\begin{tabular}{cc}
\begin{minipage}{0.5\textwidth}
  \scshape
  \footnotesize

  \centerline{\includegraphics[width=1\textwidth,trim=60 0 60 0,clip]{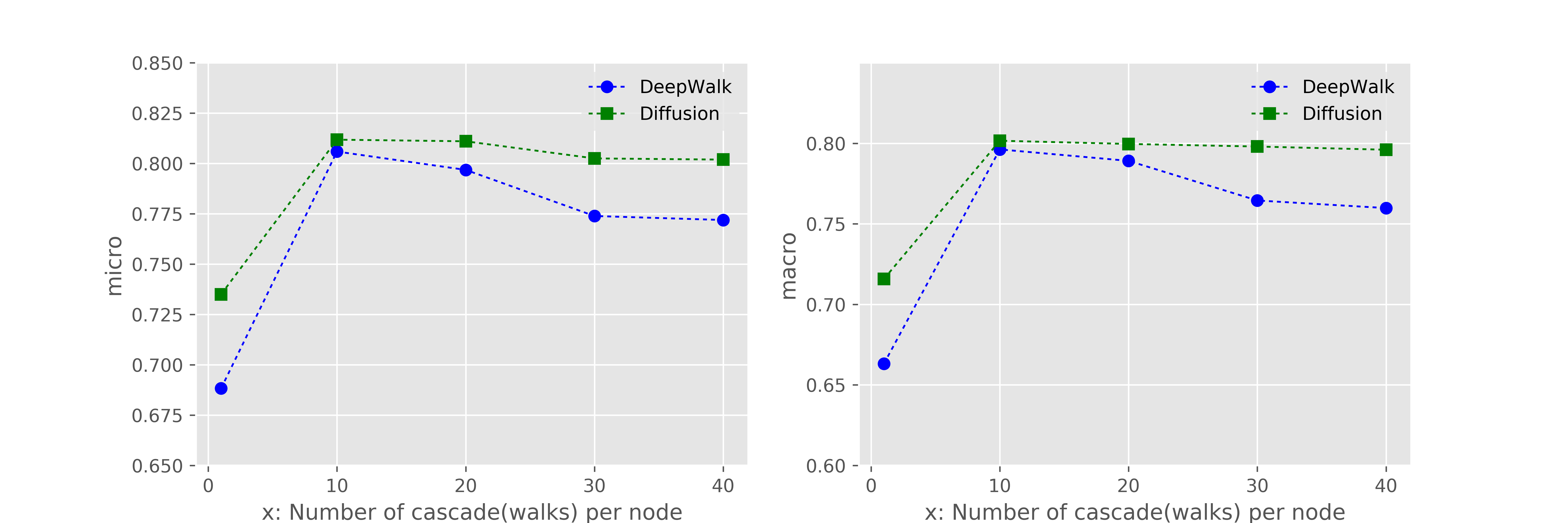}}

  \centerline{(a) Cora, $T^{c}=40(T^{w}=40)$}

\end{minipage}

\vfill

\begin{minipage}{0.5\textwidth}
  \scshape
  \footnotesize

  \centerline{\includegraphics[width=1\textwidth,trim=60 0 60 0,clip]{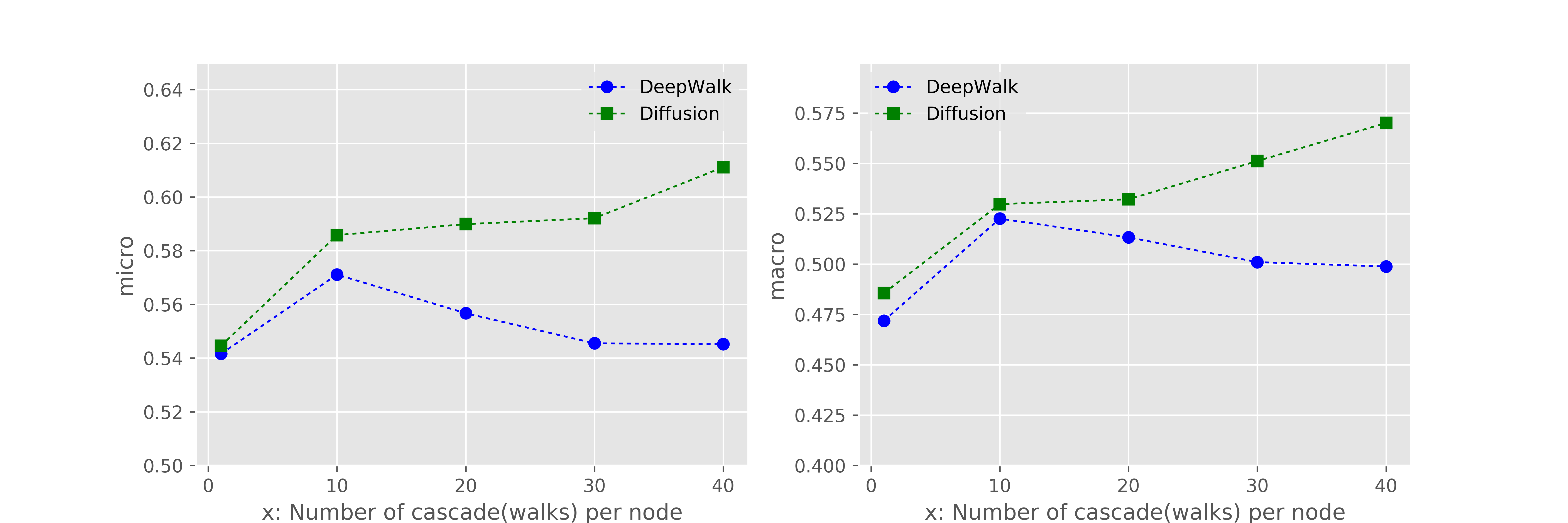}}

  \centerline{(b) Citeseer, $T^{c}=40(T^{w}=40)$}

\end{minipage}

%\end{tabular}

\caption{The comparison results of DeepWalk and Diffusion when varying the number of samplings for each node.}

\label{fig:figure_3}

\end{figure}

\begin{figure}
  \centering
  \includegraphics[width=0.5\textwidth,trim=60 0 60 0,clip]{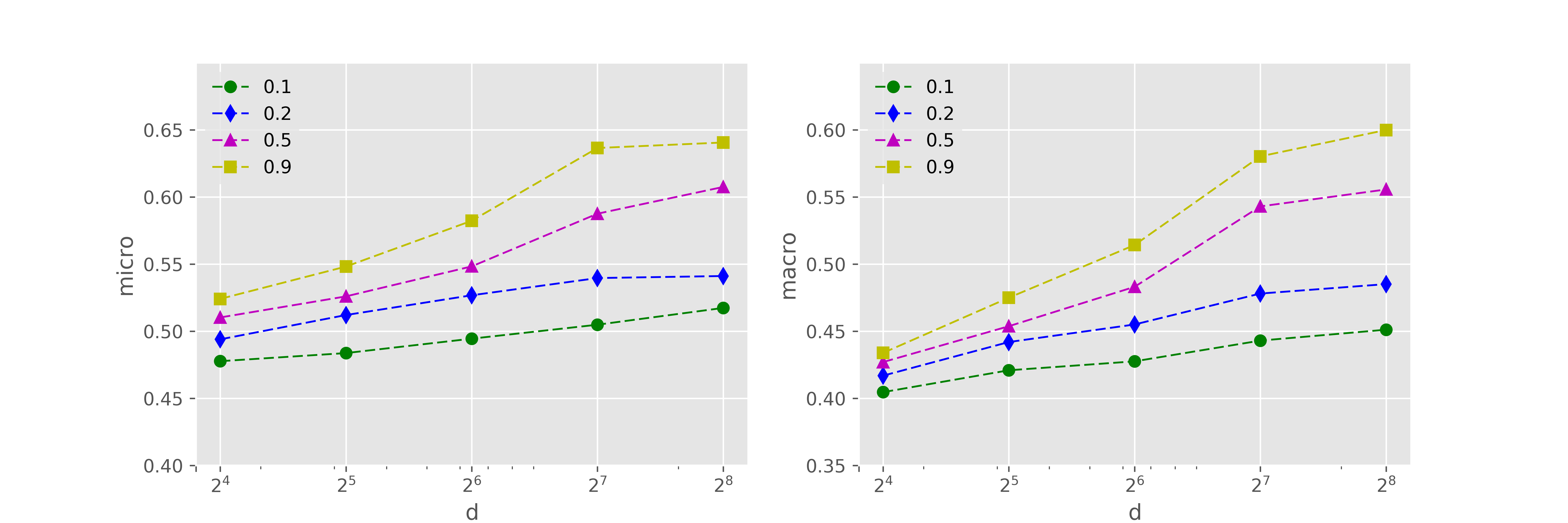}
  \caption{The classification performances on Citerseer with different representation sizes of the learned representations: $T^{c}=10,\tau=1$}
  \label{fig:figure_4}
\end{figure}

\subsection{Node Classification}

We evaluate the proposed method under the task of node classification with Wiki network, Cora network and Citeseer network. The corresponding results are reported in \reftab{table2}, \reftab{table3} and \reftab{table4} respectively. Best performances are marked in bold. Following the experiment procedure used in many network embedding literatures, we randomly sample a portion of nodes as the training set and the rest nodes as testing set. The portion is varied from 10\% to 90\%.

In order to facilitate the comparison, the embedding dimension are set to be $d=128$ for our method and baseline models. The low-dimensional results of all embedding models are trained with one-vs-rest logistic regression provided in \cite{fan2008liblinear}. After obtaining the embeddings of all models, we run the supervised training procedure ten times and calculate the average performance in terms of both Macro-F1 and Micro-F1 for each model. Here, Macro-F1 is a metric which gives weight to each class, and Micro-F1 is a metric which gives weight to each instance.

In our model, in order to grasp the network structure more precisely, all vertices in a network are treated as seeds. Specially, for each vertex, we start the diffusion $\tau$ times to enrich the training cascades.

For Deepwalk, we set the \textit{window size as 10}, \textit{number of walks for each node as 40}, \textit{walk length as 40}. For GraRep, the maximum matrix transition step $k_{max}$ is set as 4.

\subsubsection{Wiki}
From the results reported in \reftab{table2}, we can see that our method outperforms the baseline models when training rate is higher than 30\% in both Micro-F1 and Macro-F1. Moreover, The performance gain of diffusion based embedding in Micro-F1 is more prominent.

\subsubsection{Citeseer}
As indicated in \reftab{table3}, our method outperforms other baseline models in Macro-F1 score and Micro-F1 score except when training rate is 70\%. Moreover, our method achieves stable results when varying the percentage of labeled nodes. Although LINE and DeepWalk also have good results when training rate is less then 30\%, the proposed method has significant advantages when training rate exceed 80\%.

\subsubsection{Cora}
We observe from \reftab{table4} that our method achieves better performance in both Macro-F1 and Micro-F1 measures except in 10\% of training samples. The diffusion embedding perform especially well when training rate locates between 20\% to 70\% where Micro-F1 and Macro-F1 scores are at least 1\% higher than baseline models.

\subsection{A Case Study in DBLP}

\begin{figure}
  \centering
  \includegraphics[width=0.5\textwidth,trim=60 0 60 0,clip]{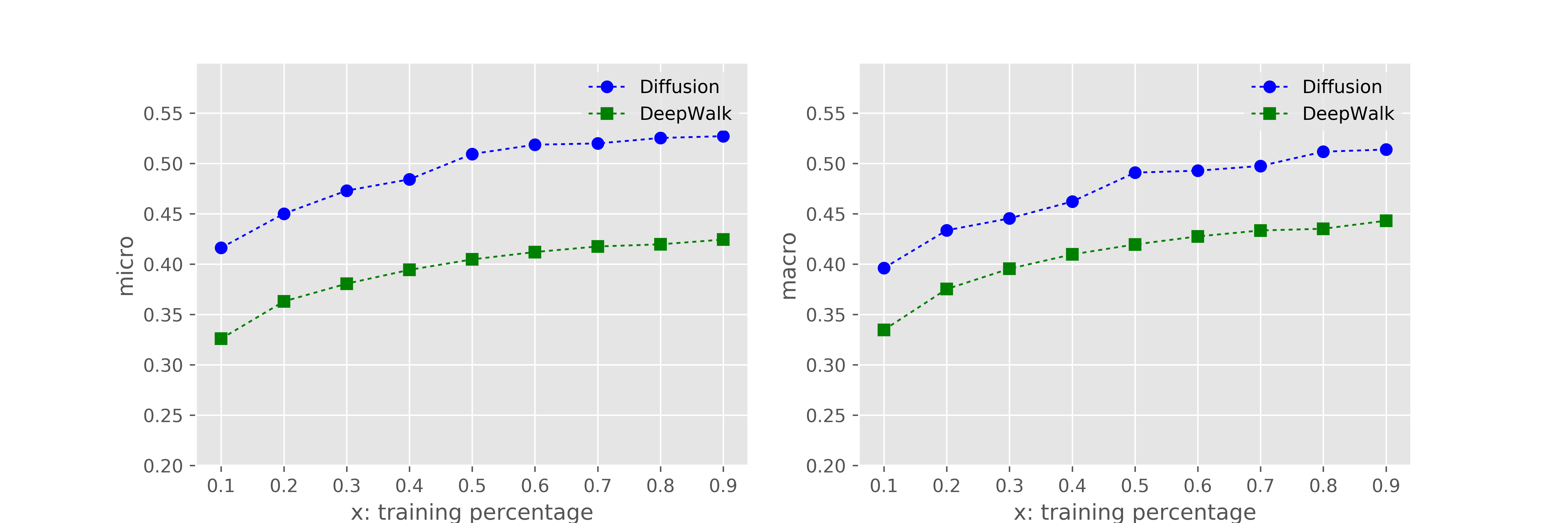}
  \caption{The classification results in highly biased DBLP network}
  \label{fig:figure_5}
\end{figure}

In this subsection, we launch a case study on DBLP data to verify the robustness of our method in the highly biased network. The overall connections of a biased network are very sparse except in a few high-degree nodes. The exceptional nodes are considered as important nodes which have much more neighbors than other nodes. The random walk in graph sampling \cite{gjoka2010walking} is easily biased towards high-degree nodes. In our experiment, we construct a biased network of DBLP in which there are only a few high-degree nodes. We compare the new proposed diffusion model with DeepWalk in classification task to see whether the biased structure will affect the performances.

The variations of Micro-F1 and Macro-F1 with the percentage of labeled nodes are plotted in \reffig{fig:figure_5}. The results show that in highly unbalanced network, the diffusion models is less affected by the biased structure comparing with DeepWalk. It is probably because that, by using more local information and global information, our model can eliminate the impact of biased structures to some extent.

\subsection{Parameters and Analysis}

In this section, we compare our method mainly with DeepWalk to analyze the parameter sensitivity of our model.

As we demonstrate in the previous sections, the time window $T^{c}$ is the time length we choose to observe in the diffusion process, the number of cascades $\tau$ for each node is related to the total amount of training corpus. Since an cascade without timestamps is much similar to random walk sequence, we bridge the correspondence of $T^{c}$ and $\tau$ with the walk length $T^{w}$ and the number of walks for each node $\gamma$ in DeepWalk respectively.

Stochastic methods need plenty of instances to simulate the true distribution. Consequently, they are more easily suffered in performance from insufficient training samples. As in network embedding, the graph sampling methods are highly depending on the number of samplings since more sequences are more likely to detect the true structures of the network.

At the very beginning, we consider an extreme case in which the number of cascades(walks) for each node is 1 and the time window(walk length) is 10. We plot the Macro-F1 and Micro-F1 for all three datasets. As illustrate in \reffig{fig:figure_2}, our model achieves better performances than DeepWalk. Even with scanty number of cascades, the diffusion based embedding is also capable to detect valid network structures. One explanation of this phenomenon is that methods such as DeepWalk focus on local structure detection and therefor requires repeated sampling on each vertex to guaranty the network structures could be captured adequately. However, the diffusion based model considers global information in the procedure of sampling and inference and therefore is more robust to the sampling numbers in each vertex.

In the second case, the time window(walk length) is set to be 40, and the number of cascades(walks) for each node is varied from 1 up to 80. We record the Macro-F1 and Micro-F1 in Cora dataset and Citeseer dataset when training set ratio is fixed as 0.8. The \reffig{fig:figure_3} show that our model is better performed than DeepWalk and achieves more stable performance when the number of cascades is larger than 10.

Finally, we plot the performances in Citeseer by varying the dimensions of the learned representations. We choose the sampling ratio $(0.1,0.2,0.5,0.9)$ as examples to report the classification results. As shown in \reffig{fig:figure_4}, the results improves significantly when dimensions is less than 128 and become stable afterwards.

\section{Conclusion}
We proposed a new embedding method to learn the low dimensional representations of network nodes. A revised sampling method based on diffusion theory is proposed to capture the network structures. The proposed method overcomes the disadvantages of traditional random walk based methods from two aspects. First, our model is less sensitive to sampling frequency on each node. Second, our method is more robust to unbalanced network structures. Experiments on the node classification task verify the effectiveness and efficiency of our method in capturing network structures. Future work can be focused on large-scale cascades data in the context of dynamic network embedding.

\bibliographystyle{ACM-Reference-Format}

\end{document}